\documentclass[letterpaper, 10 pt, conference]{ieeeconf}  
\IEEEoverridecommandlockouts                              

\usepackage{hyperref}
\usepackage{amsmath}
\usepackage{graphicx}
\usepackage{pifont,xcolor} 
\usepackage{multirow}
\usepackage{stfloats}
 \usepackage{booktabs} 
\usepackage{placeins}
\usepackage{subcaption}
\usepackage{array}
\usepackage{algorithm}
\usepackage{algorithmic}
\usepackage{xcolor}
\usepackage{hyperref}
\definecolor{slightdarkgreen}{rgb}{0,0.7,0} 

\newcommand{\cmark}{\textcolor{slightdarkgreen}{\ding{51}}}%
\newcommand{\xmark}{\textcolor{red}{\ding{55}}}

\title{\Huge
MV-UMI: A Scalable Multi-View Interface for Cross-Embodiment Learning
}

\definecolor{nyupurple}{RGB}{87, 6, 140}

\author{
{Omar Rayyan*$^1$}\thanks{*Corresponding author: olr7742@nyu.edu}\hspace{2mm}
{John Abanes$^1$}\hspace{2mm}
{Mahmoud Hafez$^1$}\hspace{2mm}
{Anthony Tzes$^1$}\hspace{2mm}
{Fares Abu-Dakka$^2$} \\[4pt]
New York University Abu Dhabi \\[4pt]
\href{https://mv-umi.github.io}{\textcolor{nyupurple}{https://mv-umi.github.io}}
\thanks{$^{1}$Center for Artificial Intelligence and Robotics (CAIR) NYU Abu Dhabi, $^{2}$Mechanical Engineering Program, NYU Abu Dhabi}%
}

\begin{document}

\maketitle
\thispagestyle{empty}
\pagestyle{empty}

\begin{abstract}

Recent advances in imitation learning have shown great promise for developing robust robot manipulation policies from demonstrations. However, this promise is contingent on the availability of diverse, high-quality datasets, which are not only challenging and costly to collect but are often constrained to a specific robot embodiment. Portable handheld grippers have recently emerged as intuitive and scalable alternatives to traditional robotic teleoperation methods for data collection. However, their reliance solely on first-person view wrist-mounted cameras often creates limitations in capturing sufficient scene contexts. In this paper, we present MV-UMI (Multi-View Universal Manipulation Interface), a framework that integrates a third-person perspective with the egocentric camera to overcome this limitation. This integration mitigates domain shifts between human demonstration and robot deployment, preserving the cross-embodiment advantages of handheld data-collection devices. Our experimental results, including an ablation study, demonstrate that our MV-UMI framework improves performance in sub-tasks requiring broad scene understanding by approximately 47\% across 3 tasks, confirming the effectiveness of our approach in expanding the range of feasible manipulation tasks that can be learned using handheld gripper systems, without compromising the cross-embodiment advantages inherent to such systems.
Videos can be found here: \href{https://mv-umi.github.io}{https://mv-umi.github.io}

\end{abstract}

\section{INTRODUCTION}

Imitation Learning (IL) provides a compelling pathway toward acquiring general robot policies capable of performing long-horizon tasks across diverse environments. This approach, particularly through supervised methods like Behavioral Cloning (BC), enables robots to acquire complex behaviors by learning to imitate human-directed actions in response to observations. Recent advances in architectures that better model this mapping \cite{chi2023diffusion, lee2024behavior, zhao2023learning, shafiullah2022behavior}, coupled with enhancements in embodiments and hardware integrations \cite{wu2024gello, zhao2025aloha}, have made this route increasingly convincing.

Recent studies on data scaling laws in imitation learning \cite{lin2025data} show that robot policy performance follows 
training scenario diversity, emphasizing the need for extensive and diverse data for robust policies. Data collection typically lies between two extremes. On one end, robot teleoperation enables the acquisition of high-quality, precise data with minimal embodiment discrepancies. However, this method is time-consuming and costly, as it requires an actively operated robot. On the other end, the internet is replete with videos of humans performing various tasks. However, substantial effort is required to establish structured explicit mappings between observed states and actions from these videos.

\begin{figure}[t]
    \centering
    \includegraphics[width=1.02\linewidth]{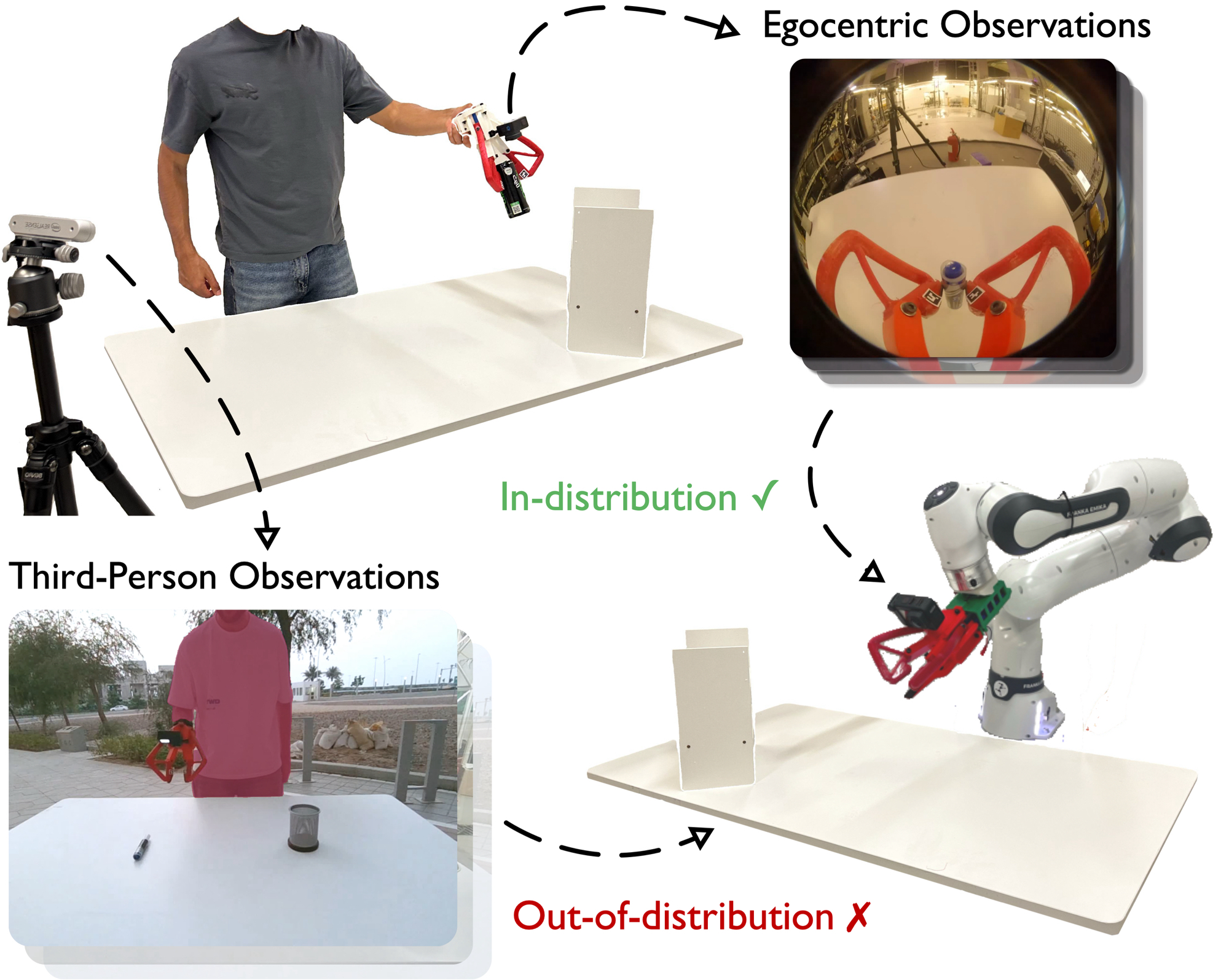}
\caption{Portable handheld gripper systems capture egocentric wrist-mounted views that remain consistent across human demonstrations and robot deployment. In contrast, third-person views are human-aligned during data collection but robot-aligned at deployment, creating an out-of-distribution gap that \textbf{MV-UMI} closes.}
    \label{fig:placeholder}
    \vspace{5pt}

\end{figure}

As a middle-ground, portable handheld grippers \cite{shafiullah2023bringing, chi2024universal, seo2025legato, wu2024fast, etukuru2024robot} have emerged as inexpensive and intuitive to use data collection devices. By relying exclusively on a wrist-mounted camera, they enable non-experts to record demonstrations without the need for a robotic manipulator. While this egocentric viewpoint minimizes visual discrepancies between training and deployment, resulting in cross-embodiment policies, it demands that the robot maintain a longer memory context to recall scene elements that move out of the constraining wrist view. 

In this work, we propose a novel framework that augments the conventional wrist-mounted camera in handheld gripper systems with a third-person camera viewpoint, without incurring distributional shifts. We achieve this by performing real-time masking of the human demonstrator in the third-person video stream, effectively removing the operator’s presence from the training data. As a result, the model benefits from a broader view of the environment, while relying less on memory for scene remembrance. A side benefit we find of this masking is its removal of correlations between the demonstrator’s motions and the gripper’s actions, encouraging the policy to focus on task-relevant cues such as the manipulated objects rather than overfitting to human-specific signals. We also utilize a custom-made three-jaw gripper for some of the tasks that require greater dexterity in this work. This design allows for greater payload weight in comparison to other hand-held devices, at the cost of its volume. Schematics and instructions to reproduce the hardware are open-sourced separately. The hardware aspect is not a key focus in this paper. 

\textbf{Summary of Contributions:}
\begin{enumerate} \item \textbf{Multi-View Cross-Embodiment Framework:} MV-UMI fuses wrist-mounted and third-person views using SAM-2 segmentation and inpainting to eliminate domain shift, boosting performance in context-dependent tasks by 47\%.
\item \textbf{End-to-End Open-Source System:} Complete pipeline, including hardware design, data collection, training code, and deployment tools, is publicly released to advance cross-embodiment manipulation research, \href{https://mv-umi.github.io}{https://mv-umi.github.io}.
\end{enumerate}

\section{Related Works}

\subsection{Robot-Free Data-Collection Devices}

Teleoperation has long been the standard approach for collecting demonstrations in robot learning, with systems ranging from general-purpose input devices such as phones and VR controllers \cite{mandlekar2018roboturk, dass2024telemoma, qin2023anyteleop,rayyan2024mujocoar} to more specialized hardware-based teacher–follower systems \cite{wu2024gello, yang2025ace, fu2024mobile}. A common limitation across these approaches, however, is the requirement of having a physical robot available during data collection. In practice, this shifts the bottleneck from the availability of human demonstrators to the availability of robots, which remain costly and limited in access.

In recent years, portable handheld grippers have gained interest as a cost-effective alternative to teleoperation for collecting manipulation demonstrations. One of the earliest explorations in this was demonstrated by kitchen-inspired tongs \cite{praveena2019characterizing}, with motion capture markers to record kinematic trajectories. Subsequent systems \cite{pari2021surprising, song2020grasping} advanced this work by incorporating a wrist-mounted camera on the device to capture a view that would align with the observation space during deployment on a robot. This allowed us to avoid reliance on motion capture markers in favor of directly extracting action-relevant features from recorded frames.

Building on these efforts, several handheld interfaces have been proposed to further scale data collection while improving data quality. In Dobb-E and RUM frameworks \cite{shafiullah2023bringing, etukuru2024robot}, a low-cost “stick” device is used with an iPhone acting as a wrist-mounted sensor, capturing both RGB-D and motion data. Meanwhile, the UMI framework \cite{chi2024universal, ha2024umi} has become the go-to device for quick data collection, owing to its hardware-agnostic design and imitation learning pipeline. It utilizes a mounted GoPro camera to maximize the field-of-view and uses visual SLAM for posture estimation. Unlike Dobb-E’s approach however, this SLAM method requires pre-collection calibration and depends on sufficiently textured environments to maintain reliable tracking of pose. Complementing these, Fast-UMI \cite{wu2024fast} simplifies UMI deployment with a wrist-mounted tracking module, while Legato \cite{seo2025legato} introduces a handheld gripper that unifies action and observation spaces across embodiments.

Beyond hand-held devices, works have explored other approaches for demonstration collection. DexCap \cite{wang2024dexcap} introduces a glove-based system that captures wrist and finger motions alongside egocentric RGB-D observations. In a complementary direction, AR2-D2 \cite{duan2023ar2} removes the need for physical robots entirely by using iPhone's AR application to overlays a virtual robot arm onto real-world scenes, allowing users to record demonstrations without the need of a physical robot.

While typical imitation learning setups \cite{zhao2023learning} and datasets \cite{khazatsky2024droid} that are collected using teleoperation use both third-person and egocentric viewpoints, current hand-held grippers rely solely on wrist-mounted cameras. This limits the ability of integrating other datasets during training, and necessitates that policies learn long temporal dependencies for objects that move out of view, particularly in multi-step tasks.

\subsection{Cross-Embodiment Learning}

\begin{table}
\vspace{5pt}
\caption{Comparison of State-of-the-Art Cross-Embodiment Frameworks}
\label{tab:cross_embodiment}
\centering
\renewcommand{\arraystretch}{1.2}
\begin{tabular}{|c|c|c|c|}
\hline
\multirow{2}{*}{\textbf{Method}} & \multicolumn{2}{c|}{\textbf{Viewpoints}} & \multirow{2}{*}{\shortstack{\textbf{Without Robot}\\\textbf{Teleoperation}}} \\ \cline{2-3}
 & \textbf{1st-Person} & \textbf{3rd-Person} & \\ \hline
Dobb-E \cite{shafiullah2023bringing} & \cmark & \xmark & \cmark \\ \hline
UMI \cite{chi2024universal} & \cmark & \xmark & \cmark \\ \hline
Fast-UMI \cite{wu2024fast} & \cmark & \xmark & \cmark \\ \hline
Legato \cite{seo2025legato} & \cmark & \xmark & \cmark \\ \hline
Shadow \cite{lepertshadow} & \xmark & \cmark & \xmark \\ \hline
Mirage \cite{chen2024mirage} & \xmark & \cmark & \xmark \\ \hline
\textbf{MV-UMI (Ours)} & \cmark & \cmark & \cmark \\ \hline
\end{tabular}

\end{table}

As we advance toward generalizable models that operate across diverse environments and platforms, enabling policies to transfer across different robot embodiments has become a critical area of exploration. The goal is to be able to transfer policies from one robot embodiment to another without the need for additional data collection.


Shadow \cite{lepertshadow} introduces a data editing technique that addresses this challenge using composite segmentation masks. During training, the source robot's pixels are masked and replaced with a rendered segmentation mask of the target robot in the corresponding end-effector pose. This alignment ensures a consistent input distribution between training and evaluation, enabling policy transfer despite visual differences between robot embodiments. However, this technique assumes prior knowledge of the target embodiment during policy training, which limits its generalizability. In Mirage \cite{chen2024mirage}, image-inpainting techniques are used to replace the source robot with the target robot during evaluation. By ``painting over'' the source robot collected data with an in-painted image of the target, Mirage minimizes visual discrepancies without introducing additional deployment latency. This approach, however, also requires prior knowledge of the target embodiment during training.

A complementary line of work \cite{bharadhwaj2024track2act, xu2024flow} explores designing intermediate interfaces that abstract away embodiment-specific visual cues. Im2Flow2Act introduces an object-flow representation that tracks the motion of manipulated objects in the scene and uses this as the primary input to a diffusion-based policy for manipulation. Our MV-UMI framework shares the same goal of minimizing embodiment bias in demonstrations but takes a complementary approach. Rather than restricting the observation to only certain task-relevant signals, we instead remove embodiment-specific elements that should not be visible, preserving the rest of the scene context.



\begin{figure*}[t!]
  \centering
  \includegraphics[width=0.95\linewidth]{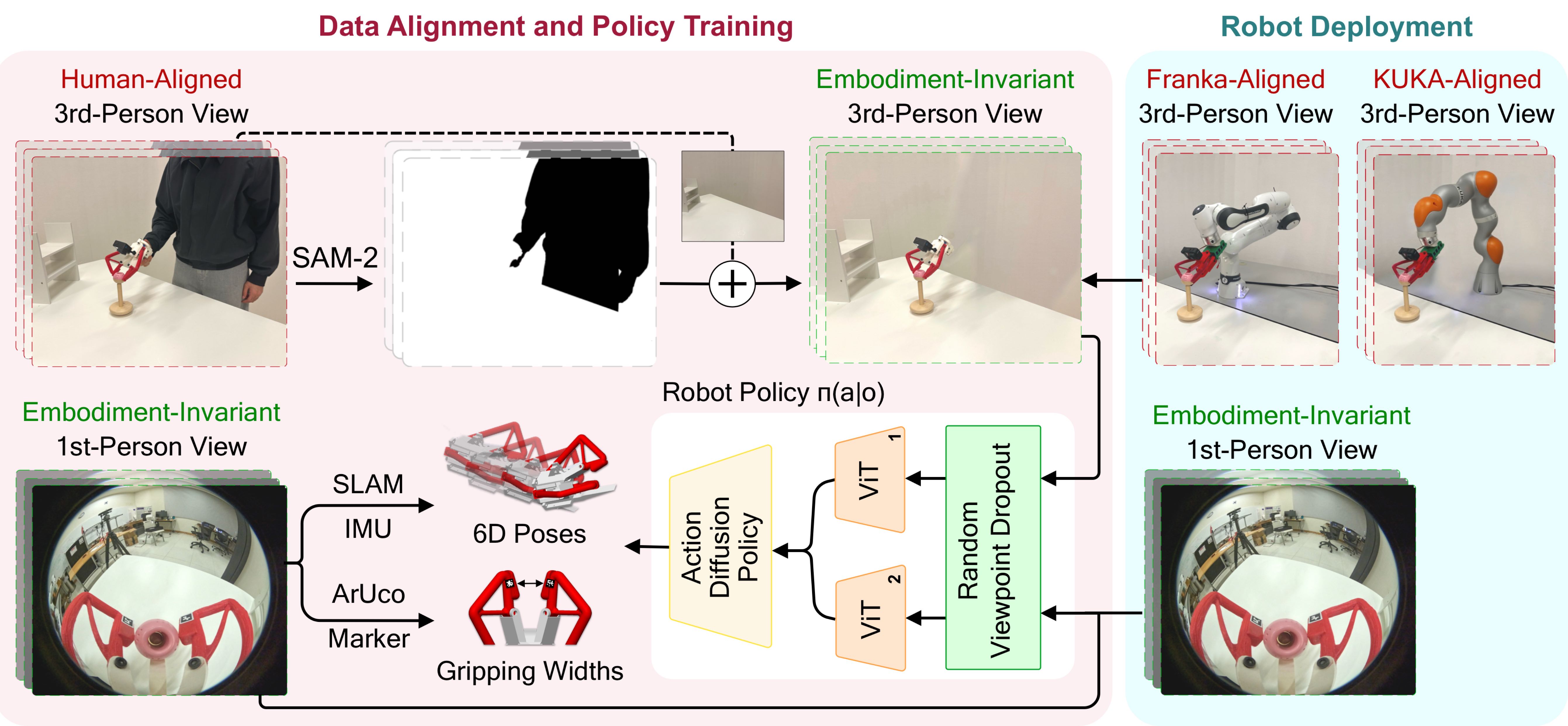}
    \caption{MV-UMI data processing and deployment pipeline. During training, human demonstrations are captured from both egocentric (wrist-mounted) and third-person cameras, with SAM-2 segmentation removing the human demonstrator and background inpainting creating robot-compatible observations. At deployment, the trained policy processes real-time multi-view inputs for cross-embodiment policy transfer.}
  \label{fig:segment}
  \vspace{-10pt}
\end{figure*}

\section{MV-UMI Framework}

\subsection{Problem Setup}
We frame our task as a \emph{cross-embodiment} learning problem in which the human demonstrator and the robot manipulator are considered distinct embodiments. Each recorded demonstration consists of:
\begin{itemize}
        \item \textbf{Egocentric camera view}: A first-person image stream, denoted \(o_t^{\text{ego}}\), recorded from the camera mounted on the handheld gripper device. This view is embodiment-invariant by nature, i.e., it remains the same in both the source and target settings.
    \item \textbf{Third-person camera view}: 
    \begin{itemize}
        \item During data collection, the overhead view is denoted as \(o_t^{\text{3rd}_{H}}\) and captures the entire scene, including the human demonstrator.
        \item During deployment, the overhead view is denoted as \(o_t^{\text{3rd}_R}\) and also captures the scene but with the robot manipulator instead.
    \end{itemize}
We do not formally calibrate the third-person camera placement across different environments. This design choice not only reduces setup effort and stays in touch with the in-the-wild data collection, but also introduces natural viewpoint variation that improves robustness during deployment.

    \item \textbf{Actions}: The handheld device pose $T$ (i.e., end-effector position/orientation), along with the gripping width $w$, recorded at each time step.
\end{itemize}


We collect trajectories of the form $
\{(s_1, a_1),\;(s_2, a_2),\;\dots,\;(s_T, a_T)\},$ where \( s_t = \bigl(o_t^\text{ego},\, o_t^{\text{3rd}_H}) \) and \( a_t = \bigl(T_t,\, w_t)\) is the handheld device pose and gripper width. Our goal is to learn a policy $\pi(a_t \mid s_t)$ that maps the state \(s_t\) to the corresponding actions \(a_t\). 


\subsection{Data Collection and Preparation \label{sec:data_collection}}

Each recording session starts with the human operator scanning a code using the wrist-camera (done per environment, not episode). This initializes shared timestamps between the egocentric and third-person view cameras. Meanwhile, the third-person camera continuously records without interruption throughout the entire session. Once the raw videos are collected, we perform several offline processing steps as described below.

\subsubsection{Segmentation}
The third-person footage depicts the human demonstrator with the handheld gripper; thus, we remove the human from each overhead frame to eliminate correlations between the human's movements and the robot's intended actions.

Specifically, we apply Segment Anything v2 ($SAM_2$) \cite{ravi2024sam} on each overhead frame as in:
\begin{equation}
    o_t^{\text{3rd}_H} : o_t^{\text{3rd-mask}} = \operatorname{SAM_2}\bigl(o_t^{\text{3rd}_H}\bigr)
    \label{eqn:sam2}
\end{equation}

This process yields a binary mask of the human region for subsequent inpainting. To prompt \textit{SAM2}, multiple points are selected from the first frame of the first episode only: positive points that indicate the person and negative points indicating the gripper to ensure proper segmentation. The keypoints of that first frame are later propagated throughout the entire session, without the need for manual prompting in each episode.  


\subsubsection{Inpainting with a Static Reference Frame}
To fill in the masked region, we use a background reference frame \(o_{\text{ref}}^{\text{bg}}\). We then in-paint the human pixels by blending the background frame into the masked area via \eqref{eqn:inpaint}:
\begin{equation}
    o_t^{\text{3rd-masked}} = \operatorname{Inpaint}\bigl(o_t^{\text{3rd}_H}, o_{\text{ref}}^{\text{bg}}, o_t^{\text{3rd-mask}}\bigr)
    \label{eqn:inpaint}
\end{equation}

This produces overhead frames that appear as if no human was present, thereby mitigating distribution shift when transferring to a robot that sees only itself and the environment. As we show in our ablation study (Section~\ref{sec:system_evaluation}), this human-removal step is critical for enabling successful policy transfer, with policies trained on unsegmented data not working robustly.

\subsubsection{Time Synchronization}
Although we initialize the camera streams via a code scan at the start, minor frame-to-frame offsets can still occur. Thus, each egocentric frame \(o_t^{\text{ego}}\) is matched to the processed overhead frame in the closest time \(o_{\tau(t)}^{\text{3rd-masked}}\). This pairing ensures that both views capture the same scene context. We then store $s_t$, defined in \eqref{eqn:pair} along with the corresponding action $a_t$:
\begin{equation}
    s_t = \Bigl(o_t^{\text{ego}},\; o_{\tau(t)}^{\text{3rd-masked}}\Bigr)
    \label{eqn:pair}
\end{equation}

\subsubsection{Action Extraction}
In a similar fashion to the UMI \cite{chi2024universal}, we extract actions directly from the egocentric \texttt{.MP4} footage. In particular, visual SLAM using  ORB-SLAM3 \cite{campos2021orb} is used together with GoPro IMU data to estimate the pose of the handheld device in each frame relative to the starting pose. The gripping width is determined by measuring the distance between two ArUco markers attached to the jaws of the handheld gripper. These pose and gripper-width estimates form the final action labels \(\{a_1, \dots, a_T\}\).

\begin{figure*}[t]
\vspace{10pt}
    \centering
    \begin{subfigure}[c]{0.19\linewidth}
        \includegraphics[width=\linewidth, trim=200 0 200 0, clip]{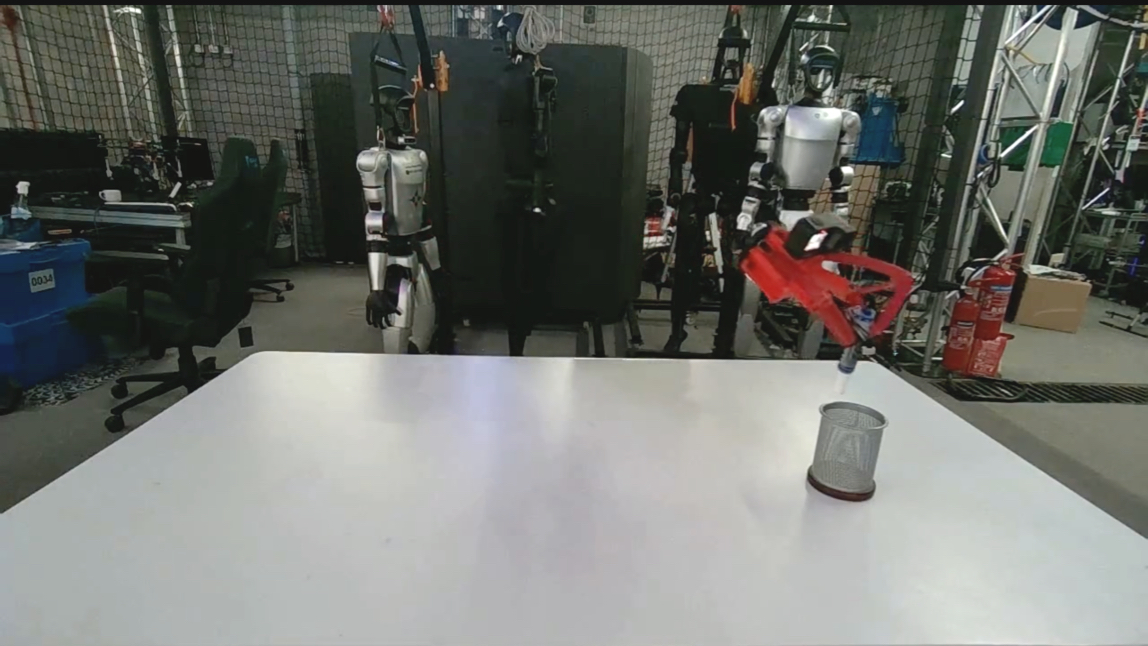}
    \end{subfigure}
    \begin{subfigure}[c]{0.19\linewidth}
        \includegraphics[width=\linewidth, trim=200 0 200 0, clip]{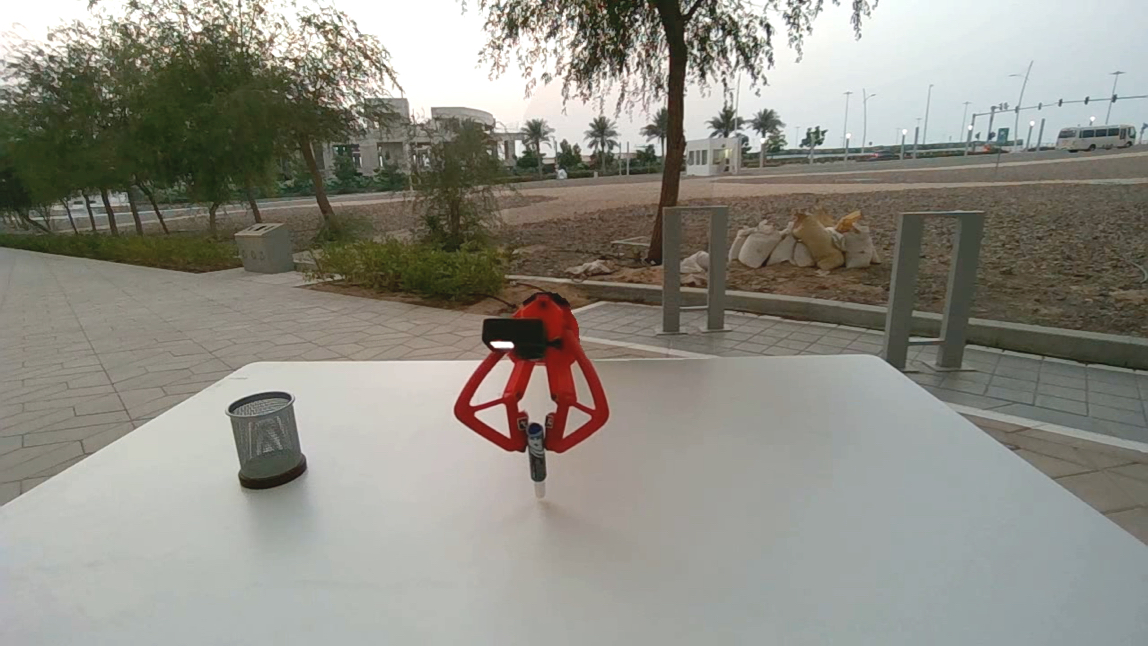}
    \end{subfigure}
    \begin{subfigure}[c]{0.19\linewidth}
        \includegraphics[width=\linewidth, trim=200 0 200 0, clip]{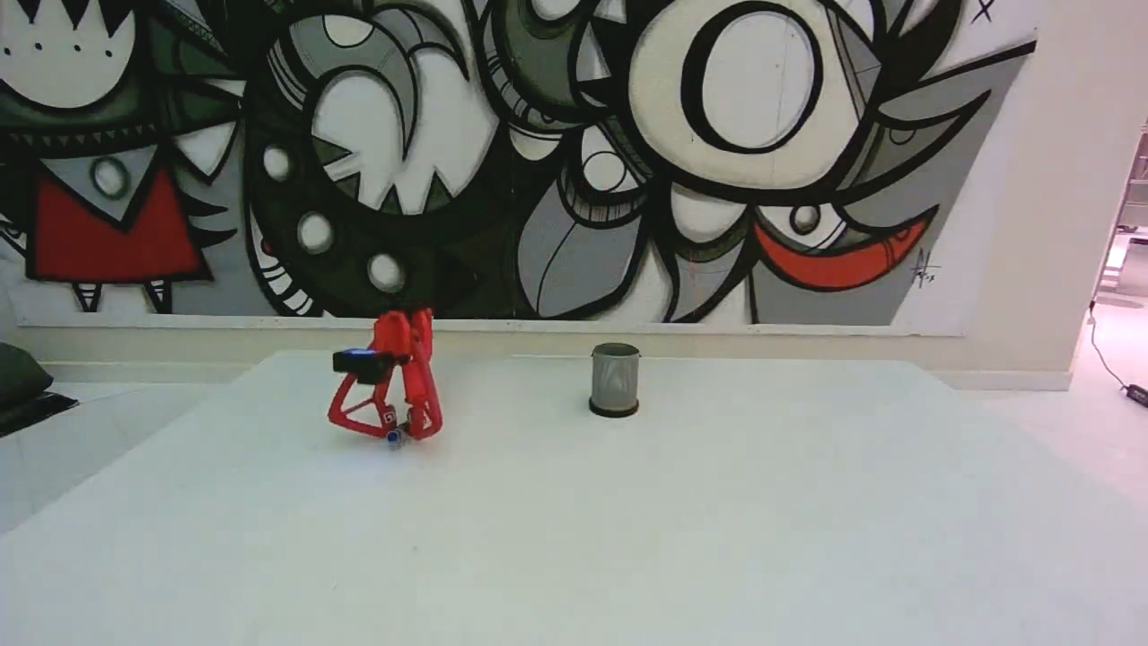}
    \end{subfigure}
    \begin{subfigure}[c]{0.19\linewidth}
        \includegraphics[width=\linewidth, trim=200 0 200 0, clip]{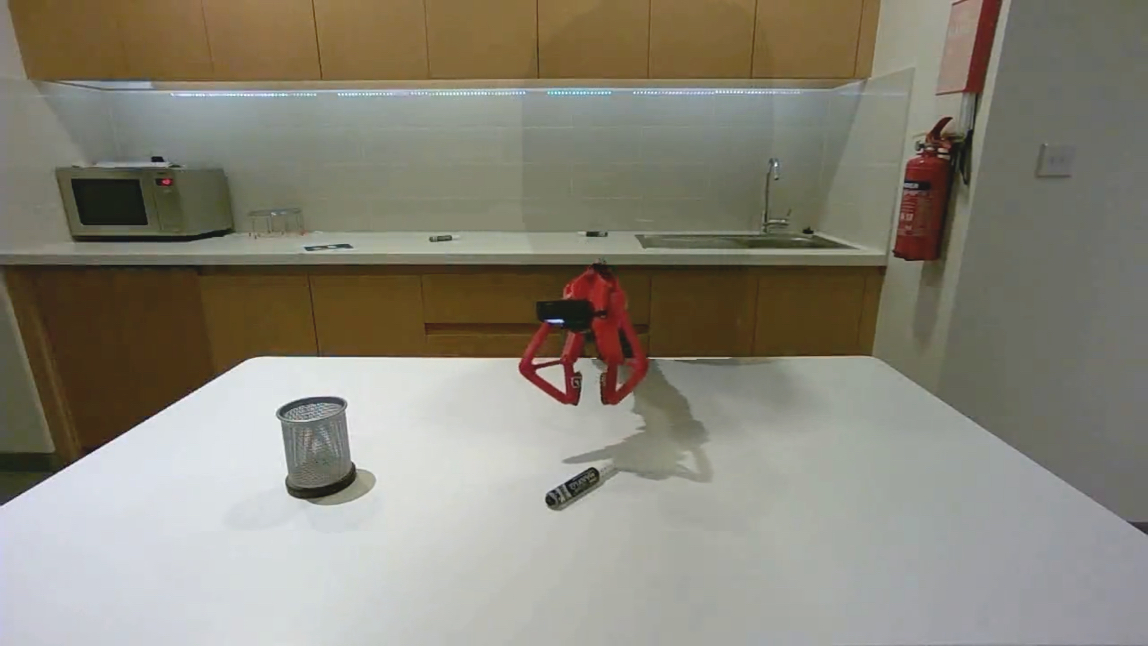}
    \end{subfigure}
    \begin{subfigure}[c]{0.19\linewidth}
        \includegraphics[width=\linewidth, trim=200 0 200 0, clip]{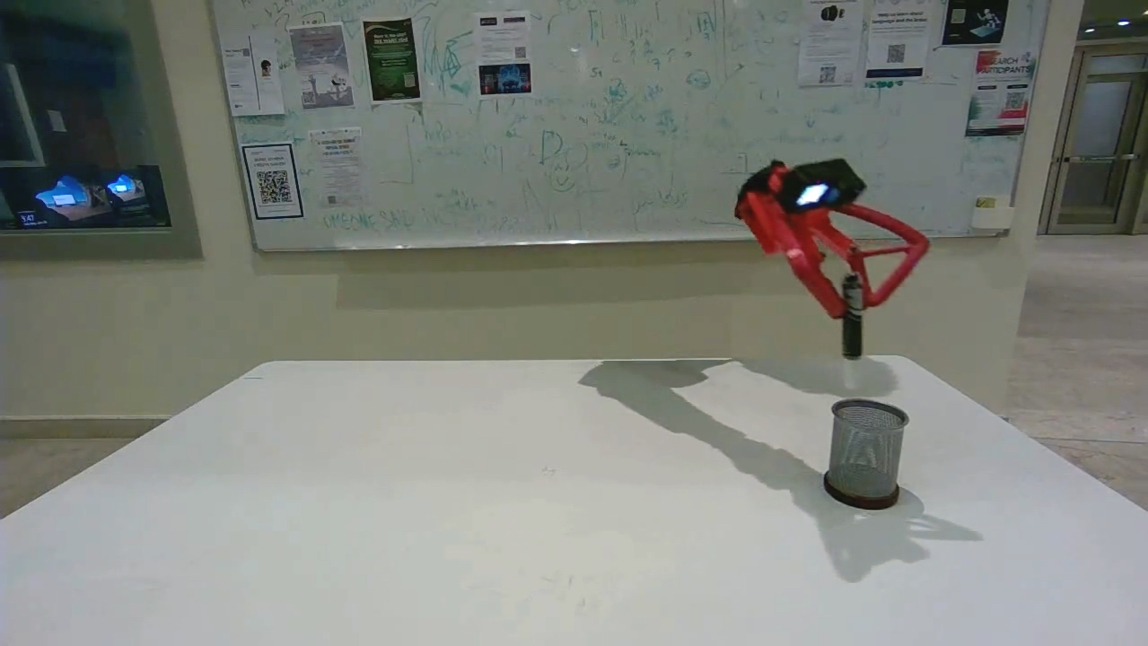}
    \end{subfigure}

    \caption{Portion of environments used with MV-UMI demonstrating robustness in \textit{in-the-wild} scenes.}
    \label{fig:in_the_wild}
    \vspace{-15pt}
\end{figure*}

\subsection{Policy Training \label{sec:training}}

We parameterize our policy \(\pi\) via an action diffusion CNN-based architecture that predicts multi-step trajectories of end-effector poses and gripping widths. Rather than outputting a single action vector at each time step, the diffusion model infers a short-horizon sequence.

\begin{figure}[h]
\vspace{5pt}
  \centering
  \includegraphics[width=1.0\linewidth]{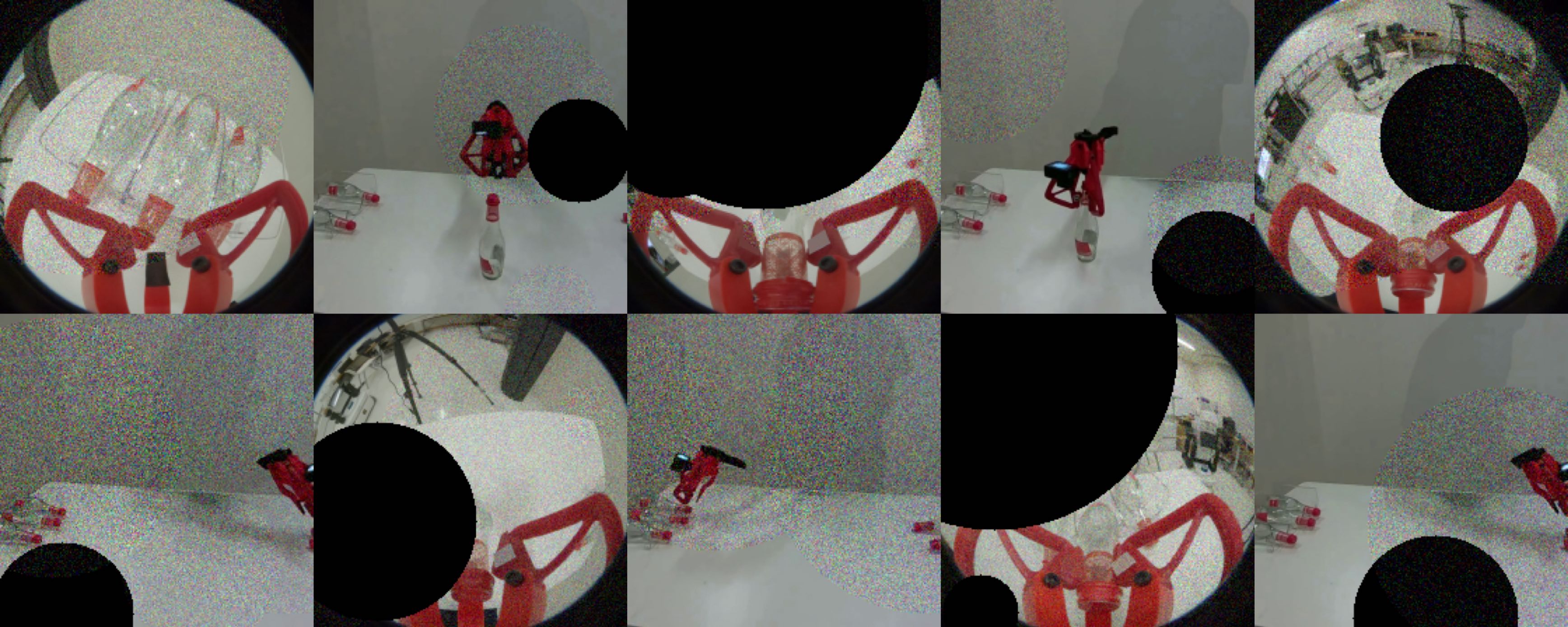}
\caption{Random viewpoint dropout augmentation used during training. We randomly mask either the egocentric or third-person camera views with varying-sized noise patches.}  \label{fig:random_dropout}
\end{figure}

To improve robustness against potential artifacts such as imperfect inpainting and occlusions, we incorporate a random dropout strategy during training. This dropout probability $p$ is exponentially reduced throughout the training, which encourages the network to avoid overfitting to a particular viewpoint or pattern. Random viewpoint dropouts and noise patches are also performed to make the model more robust against occlusions and incomplete observations, as shown in Fig.~\ref{fig:random_dropout}.

\begin{algorithm}[h]
\caption{Policy Training with Random Dropout}
\label{alg:training}
\begin{algorithmic}[1]
\STATE Initialize parameters $\theta$, dropout rate $p_0$, decay rate $\lambda$
\FOR{each iteration $t = 1, \dots, T$}
    \STATE Sample raw frames $\{o_t^{\text{ego}}, o_t^{\text{3rd}_H}\}$ and ground-truth action $a_t$
    \STATE Apply SAM2 on $o_t^{\text{3rd}_H}$ to obtain $o_t^{\text{3rd-masked}}$
    \IF{with probability $(1-p)$}
        \STATE Input both $o_t^{\text{ego}}$ and $o_t^{\text{3rd-masked}}$ to the policy
    \ELSE
        \STATE Randomly add noise patches to one view
    \ENDIF
    \STATE Compute predicted action: $\hat{a}_t \gets \pi_\theta(\cdot)$
    \STATE Compute loss: $\mathcal{L} = \operatorname{Loss}(\hat{a}_t, a_t)$
    \STATE Step policy parameters: $\theta \gets \theta - \nabla_\theta \mathcal{L}$
    \STATE Decay dropout rate: $p \gets p_0 e^{-\lambda t}$
\ENDFOR
\end{algorithmic}
\end{algorithm}

\subsection{Deployment \label{sec:deployment}}
During deployment, we study and support two configurations for handling the third-person input:
\begin{enumerate}
    \item \textbf{With robot segmentation and inpainting:} 
    Each overhead frame $o_t^{\text{3rd}_R}$ is first passed through the segmentation module 
    $o_t^{\text{3rd-mask}} = \operatorname{SAM_2}(o_t^{\text{3rd}_R})$ 
    to identify the robot manipulator. This mask is then used by the inpainting module to blend in the static background reference:
    \begin{equation}
        o_t^{\text{3rd-masked}} = \operatorname{Inpaint}\bigl(o_t^{\text{3rd}_R}, o_{\text{ref}}^{\text{bg}}, o_t^{\text{3rd-mask}}\bigr).
        \label{eqn:inpainting_blend}
    \end{equation}
    This produces input observations that are visually consistent with the data distribution used during the training. Since \textit{SAM2} operates at around 40 FPS on an A100, significantly faster than the action diffusion policy we train, the inpainting process incurs minimal computational overhead during deployment, and the real-time performance of the overall system is maintained.

    \item \textbf{Without robot segmentation:} 
    The raw third-person frame $o_t^{\text{3rd}_R}$ is directly provided to the policy without additional processing.
\end{enumerate}

We compare the effectiveness of these two deployment modes in Section~\ref{sec:system_evaluation}.

\begin{figure*}[!b]
\vspace{-5pt}
    \centering
    \begin{subfigure}[c]{0.32\linewidth}
        \centering
        \includegraphics[width=\linewidth]{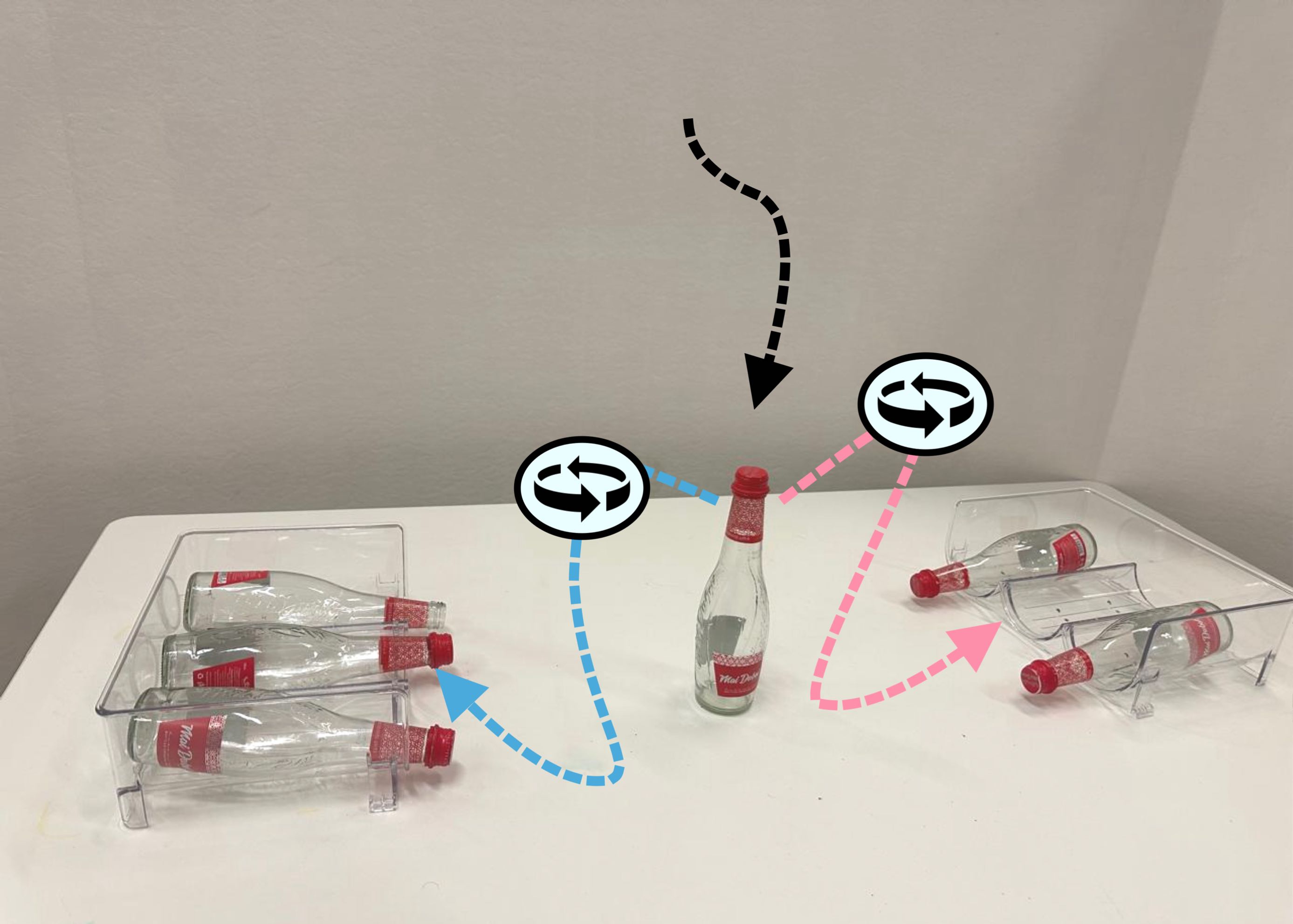}
        \caption{\texttt{Bottles-Rack-Inserter:} Insert a glass bottle into the one rack with an empty slot.}
    \end{subfigure}
    \hfill
    \begin{subfigure}[c]{0.32\linewidth}
        \centering
        \includegraphics[width=\linewidth]{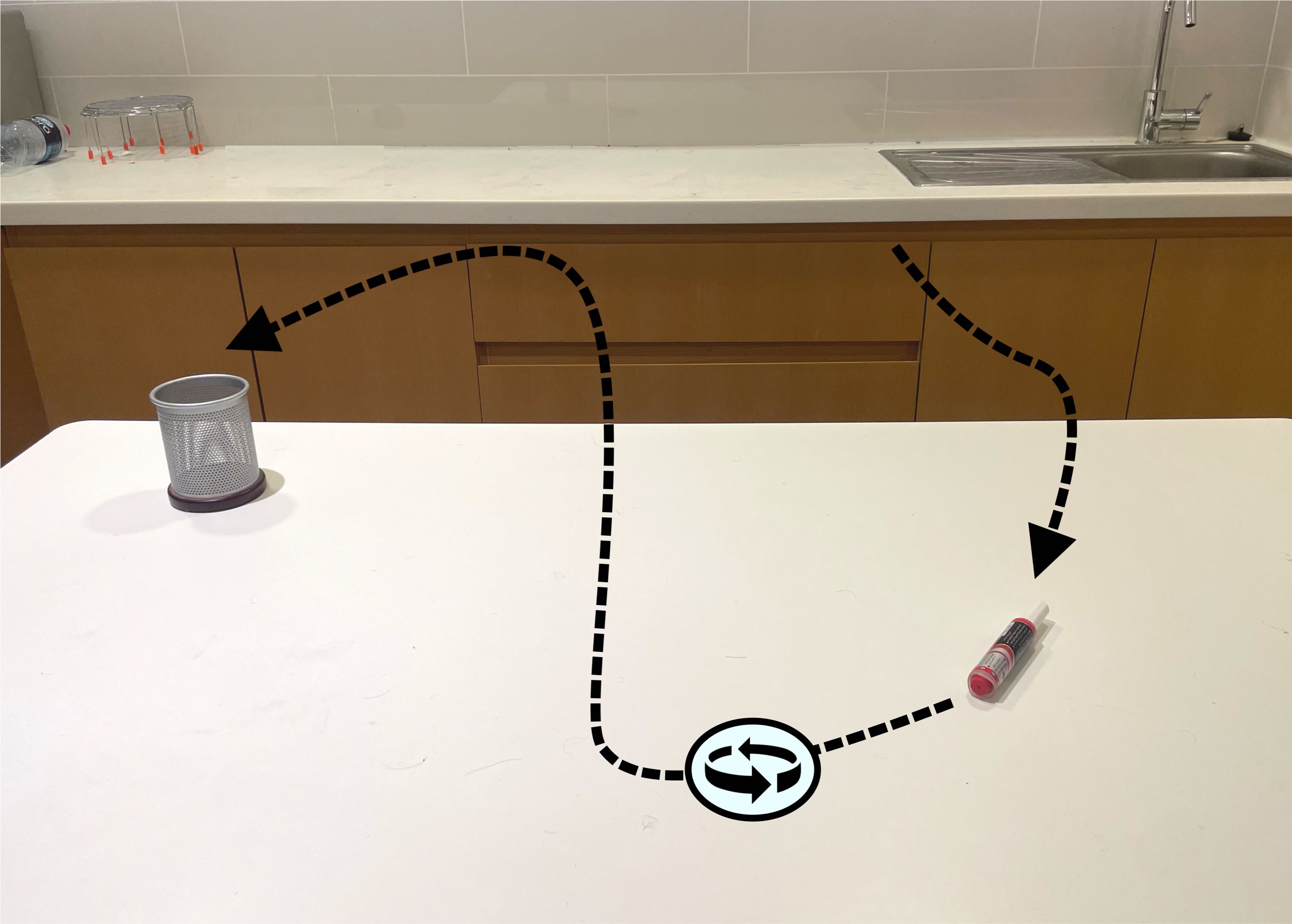}
        \caption{\texttt{Marker-Cup-Placer:} Pick and place a marker into a cup around the table.}
    \end{subfigure}
    \hfill
    \begin{subfigure}[c]{0.32\linewidth}
        \centering
        \includegraphics[width=\linewidth]{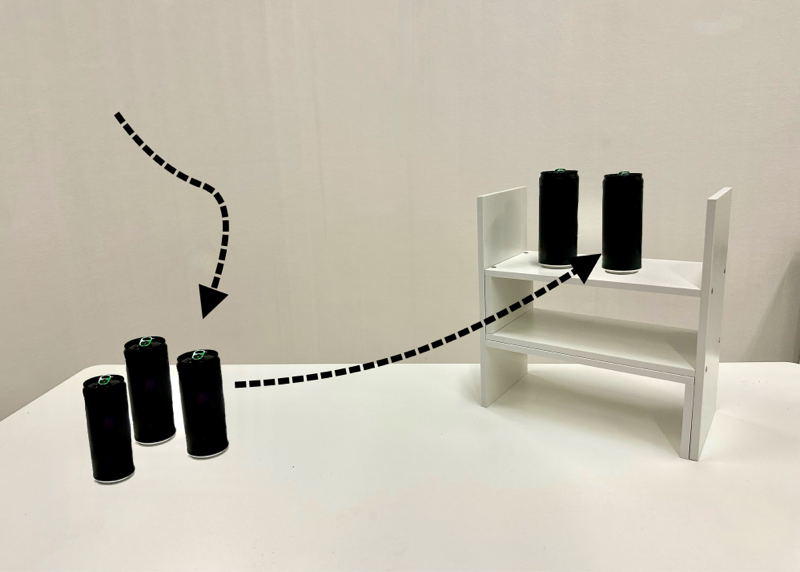}
        \caption{\texttt{Cans-Shelf-Placer:} Place a can on a shelf sitting on the other side of a table.}
    \end{subfigure}
    \caption{Task descriptions in various environments for MV-UMI system.}
    \label{fig:task_table}
\end{figure*}

\subsection{Proposed Three-Jawed gripper}

\begin{figure}[h]
\vspace{7pt}
    \centering
    \includegraphics[width=0.95\linewidth]{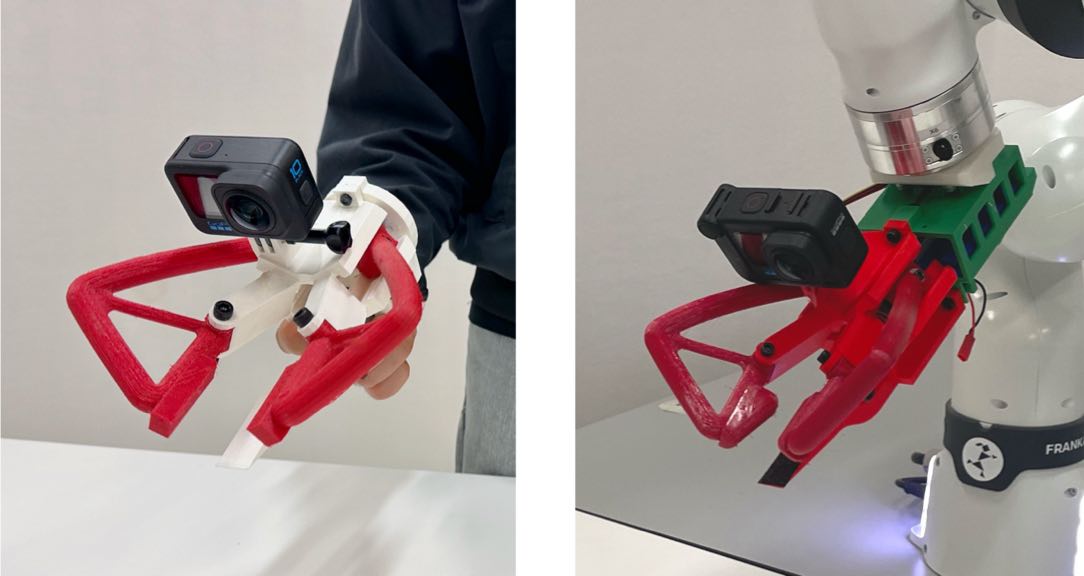}
    \caption{Our custom three-jaw gripper system includes both a 
\textbf{Left:} handheld mode and a \textbf{Right:} motorized mode. 
Each jaw is about 189 mm $\times$ 94 mm in footprint size and is printed with variable infill such that most of the jaw is soft at 25\% infill, except for the boxed areas. 
The gripping portion uses 5\% infill to maximize grip and the portions attached to fixtures use a higher infill of 45\% for stability.}

    \label{fig:three_jaw}
\end{figure}

We found the UMI gripper design to be challenging for tasks that required locking an object’s rotation. To address this, we designed a three-jaw gripper (Fig. \ref{fig:three_jaw}) that uses two compliant jaws printed in TPU 95A and a rigid third jaw. Initially, all three jaws were designed to be flexible, but this required a much larger actuation force and resulted in unreliable performance. Adding a rigid jaw mitigated both issues. The usage of soft, compliant jaws removed the need for springs. For integration with a robotic manipulator, the gripping mechanism is actuated by a linear motor. The design allows the gripper to carry payloads up to three times its own weight.

\section{System Evaluation \label{sec:system_evaluation}}

We evaluate the MV-UMI system across three key questions:
\begin{enumerate}
    \item How effective is the third-person view integration for improving policy performance in tasks requiring context beyond the egocentric camera?
    \item How important are the individual components of our data processing pipeline, such as viewpoint masking and inpainting, during training?
    \item How necessary is inpainting at inference for maintaining performance during deployment, and can the policy operate effectively without it?
\end{enumerate}

The tasks in Fig.~\ref{fig:task_table} were selected to highlight scenarios where the egocentric view alone may be insufficient. Each task benefits from broader scene understanding through the third-person view and collectively involves challenges such as object rotations and payload handling. Specifically, the \texttt{Marker-Cup-Placer} task requires awareness of distant containers that are often outside the egocentric field of view at the start, as well as precise placement once the target cup leaves the frame. 

The \texttt{Bottles-Rack-Inserter} task demands both dexterous alignment and payload handling capabilities that the UMI gripper design is unable to accomplish. One rack slot is left empty at random, requiring the model to locate it without egocentric visibility during the trajectory. Finally, the \texttt{Cans-Shelf-Placer} task emphasizes wide-scene context, as the shelf is moved to be outside the wrist camera’s view for most of the trajectory and on varying far positions on the table. In practice, we find this task challenging because the egocentric viewpoint goes out of distribution mid-execution in the changing room setting and is often occluded by the picked can.

\subsubsection{Experimental Setup}
A GoPro Hero-10 was used as the wrist-mounted gripper camera, and an Intel RealSense D455 was used to provide the third-person view. The third-person camera recorded continuously, while the episodes' beginnings and ends were determined by the timestamps of the GoPro camera videos. Scenes from our data-collection process can be seen in Fig.~\ref{fig:in_the_wild}. The third-person footage was processed as detailed in Section \ref{sec:deployment}.

\subsubsection{Ablation Study}
In our ablation study, we found that the policy trained on human-aligned data failed to complete any task, despite often moving in sensible directions. We attribute this failure to the distribution shift in the observation space between training and deployment. By contrast, the policy trained on segmented data generalized well, even when the robot was not segmented at inference. This suggests that removing the human prevents the model from overfitting to spurious correlations between human motion and gripper actions, encouraging reliance on task-relevant scene features instead. Results from this study are summarized in Table~\ref{tab:ablation_study}.

\begin{table}[h]
\vspace{5pt}
    \caption{Ablation study (relative to MV-UMI success rate).}
    \label{tab:ablation_study}
    \centering
    \small
    \renewcommand{\arraystretch}{1.1} 
    \begin{tabular}{p{0.5\linewidth}|c}
    \textbf{Ablation} & \textbf{Relative Performance} \\
    \hline
    Without Human Segmentation & 0.10 \\
    Without 3rd-Person View & 0.60 \\
    Without Robot Inpainting & 0.00 \\
    Without Robot Segmentation & 0.80 \\
    \end{tabular}
    
\end{table}

\subsubsection{Attention Map Analysis}

\begin{figure*}[t]
\centering
\includegraphics[width=1.0\linewidth]{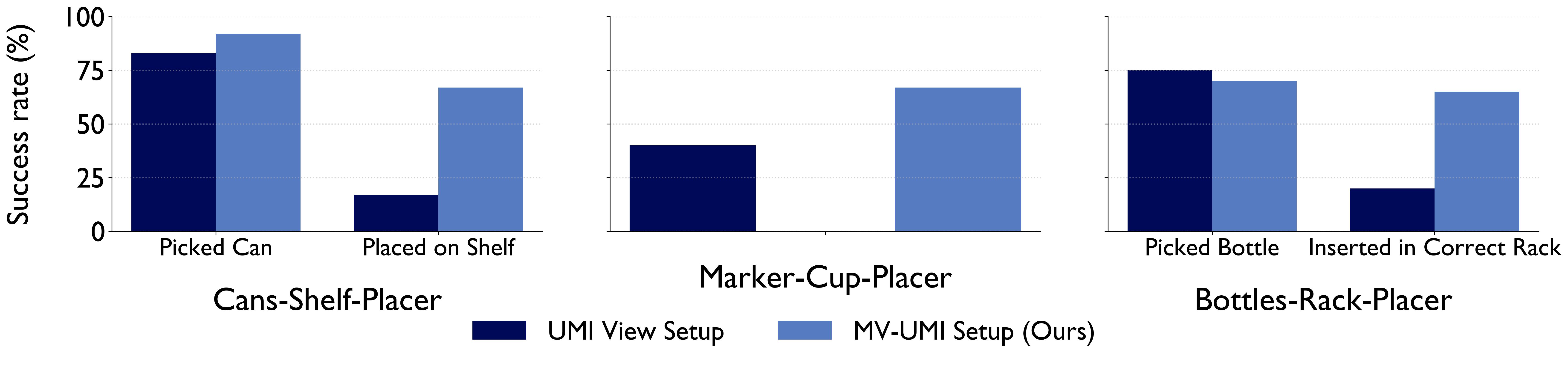} \\
\caption{Comparison of task success rates between \textit{UMI View Setup} and our proposed \textit{MV-UMI Setup}.}
\label{fig:success_rate}
\vspace{-10pt}
\end{figure*}

To better understand the source of these differences, we analyzed attention maps from the Vision Transformer (ViT) encoder under both segmented and unsegmented training conditions. Using a forward hook on the multi-head self-attention layers, we captured the query--key attention scores for the class token, which indicate how strongly the global representation attends to individual image patches. These scores were reshaped into a spatial grid, normalized, and overlaid on the original frames to visualize regions of focus.

\begin{figure}[h]
    \centering
    \includegraphics[width=0.42\linewidth]{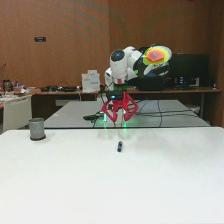}
    \includegraphics[width=0.42\linewidth]{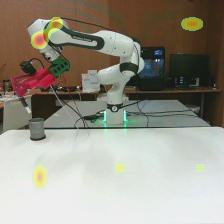} \\
    
    \small \textbf{(a)} Unsegmented model attending to embodiment and background features. \\[2mm]
    \includegraphics[width=0.42\linewidth]{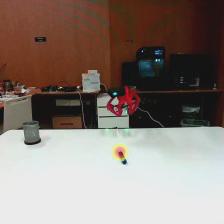}
    \includegraphics[width=0.42\linewidth]{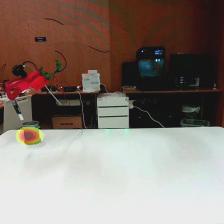} \\
    
    \small \textbf{(b)} MV-UMI model focusing on the object manipulated when the robot is segmented during runtime. \\[2mm]
    \includegraphics[width=0.42\linewidth]{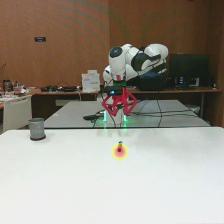}
    \includegraphics[width=0.42\linewidth]{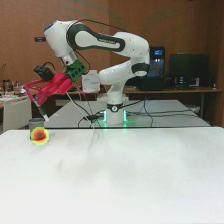} \\
    
    \small \textbf{(c)} MV-UMI model focusing on the object manipulated even when the robot is not segmented during runtime

    \caption{Attention maps from our ViT vision encoders under different training and inference conditions. }
    \label{fig:attention_maps}
\end{figure}

The visualizations (Fig.~\ref{fig:attention_maps}) directly support our ablation findings: models trained without human segmentation frequently direct attention toward embodiment-specific features such as the operator’s hand or robot arm, explaining their poor generalization. In contrast, our MV-UMI model consistently allocates attention to objects involved in the task, highlighting how data segmentation and inpainting guide the policy toward more object-centric manipulation. Interestingly, we also observe that the segmented model avoids attending to embodiment cues when the third-person view is left unsegmented during inference. We attribute this effect to the strong correlation between the human demonstrator’s motions and the corresponding robot actions; by removing the human, the model is encouraged to focus on the true task-relevant signals.


\subsubsection{Task Results}

Fig.~\ref{fig:success_rate} shows the success rates of our multi-view setup compared to the single-camera UMI baseline. The egocentric wrist-mounted camera was sufficient in initial phases of tasks when the manipulated objects remained visible, but success rates drop sharply once the target leaves the field of view or occludes it after being picked. In contrast, our MV-UMI setup improves performance in these later stages.

For example, in the \texttt{Bottles-Rack-Inserter} task, the wrist-only setup typically succeeded in the picking phase but frequently failed during placement, as it lacked awareness of which side contained the empty slot. Similar, and in the \texttt{Marker-Cup-Placer} task, failures occurred when the cup moved outside the egocentric field of view, whereas MV-UMI maintained awareness of its position. Finally, in the \texttt{Cans-Shelf-Placer} task, the single-camera setup struggled when the shelf was positioned out of view for most of the trajectory.

\subsubsection{Limitations}
While the inclusion of a third-person view enhances scene understanding and improves task performance, it also introduces some challenges. First, occlusions caused either by the human demonstrator during training or by the robot during deployment can result in incomplete segmentation under the first deployment configuration described in Section~\ref{sec:deployment}. However, we find that the second deployment configuration mitigates this issue during inference.

\section{Conclusion}
In this paper, we introduced a data collection framework that addresses a significant limitation in current handheld demonstration systems: the restricted perspective of wrist-mounted cameras. Our multi-view approach integrates third-person camera footage with egocentric observations while mitigating domain shift through real-time person-masking and inpainting. This enables policies to gain broader contextual awareness, avoid spurious human-specific signals, and maintain cross-embodiment compatibility between demonstration and deployment.


\section{Acknowledgments}

This work was performed in the Kinesis Lab, Core Technology Platform (CTP) facility of NYUAD. This work was partially supported by the NYUAD Center for Artificial Intelligence and Robotics (CAIR), funded by Tamkeen under the NYUAD Research Institute Award
CG010.

The team thanks Jorge Montalvo Navarette of NYUAD's Advanced Manufacturing Workshop for his guidance in the fabrication of several project components. We also thank Dr. Nikolaos Evangeliou of NYUAD's Robotics and Intelligent Systems Control Laboratory for his support with the electrical configuration and integration of the project's sensors and actuators.
\bibliographystyle{IEEEtran}
\bibliography{root}

\end{document}